\documentclass[runningheads]{llncs}
\usepackage{graphicx}
\usepackage{amsmath,amssymb} 
\usepackage{color}
\usepackage{verbatimbox}
\usepackage[width=122mm,left=12mm,paperwidth=146mm,height=193mm,top=12mm,paperheight=217mm]{geometry}
\begin{document}
\pagestyle{headings}
\mainmatter

\title{SPICE: Semantic Propositional Image Caption Evaluation} 

\titlerunning{SPICE: Semantic Propositional Image Caption Evaluation}

\authorrunning{Peter Anderson, Basura Fernando, Mark Johnson, Stephen Gould}

\author{Peter Anderson\textsuperscript{1}, Basura Fernando\textsuperscript{1}, Mark Johnson\textsuperscript{2}, Stephen Gould\textsuperscript{1}}


\institute{
	\textsuperscript{1}The Australian National University, Canberra, Australia\\
	{\tt\small firstname.lastname@anu.edu.au}\\
	\textsuperscript{2}Macquarie University, Sydney, Australia\\
	{\tt\small mark.johnson@mq.edu.au}\\
}

\maketitle

\begin{abstract}
	There is considerable interest in the task of automatically generating image captions. However, evaluation is challenging. Existing automatic evaluation metrics are primarily sensitive to n-gram overlap, which is neither necessary nor sufficient for the task of simulating human judgment. We hypothesize that semantic propositional content is an important component of human caption evaluation, and propose a new automated caption evaluation metric defined over scene graphs coined \emph{SPICE}. Extensive evaluations across a range of models and datasets indicate that SPICE captures human judgments over model-generated captions better than other automatic metrics (e.g., system-level correlation of 0.88 with human judgments on the MS COCO dataset, versus 0.43 for CIDEr and 0.53 for METEOR). Furthermore, SPICE can answer questions such as \emph{which caption-generator best understands colors?} and \emph{can caption-generators count?}
\end{abstract}

\section{Introduction}

Recently there has been considerable interest in joint visual and linguistic problems, such as the task of automatically generating image captions \cite{Donahue2015,Xu2015}. Interest has been driven in part by the development of new and larger benchmark datasets such as Flickr 8K \cite{Hodosh2013}, Flickr 30K \cite{Young2014} and MS COCO \cite{Lin2014}. However, while new datasets often spur considerable innovation---as has been the case with the MS COCO Captioning Challenge \cite{Chen2015}---benchmark datasets also require fast, accurate and inexpensive evaluation metrics to encourage rapid progress. Unfortunately, existing metrics have proven to be inadequate substitutes for human judgment in the task of  evaluating image captions \cite{Kulkarni2013,Hodosh2013,Elliott2014}. As such, there is an urgent need to develop new automated evaluation metrics for this task \cite{Elliott2014,Bernardi2016}. In this paper, we present a novel automatic image caption evaluation metric that measures the quality of generated captions by analyzing their semantic content. Our method closely resembles human judgment while offering the additional advantage that the performance of any model can be analyzed in greater detail than with other automated metrics.

\begin{figure}[t]
	\vskip 0.2in
	\begin{center}
		\centerline{\includegraphics[width=1.0\columnwidth]{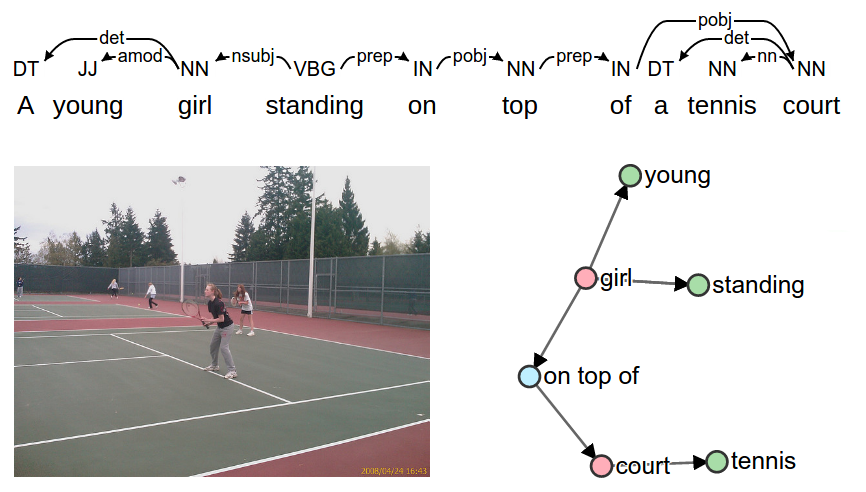}}
		\caption{Illustrates our method's main principle which uses semantic propositional content to assess the quality of image captions. Reference and candidate captions are mapped through dependency parse trees (top) to semantic \textit{scene graphs} (right)---encoding the objects (red), attributes (green), and relations (blue) present. Caption quality is determined using an F-score calculated over tuples in the candidate and reference scene graphs			
		}
		\label{fig:motivation}
	\end{center}
	\vskip -0.2in
\end{figure}

One of the problems with using metrics such as Bleu \cite{Papineni2002}, ROUGE \cite{Lin2004}, CIDEr \cite{Vedantam2015} or METEOR \cite{denkowski2014} to evaluate captions, is that these metrics are primarily sensitive to n-gram overlap. However, \textit{n-gram overlap is neither necessary nor sufficient for two sentences to convey the same meaning} \cite{Gimenez2007}. 

To illustrate the limitations of n-gram comparisons, consider the following two captions (a,b) from the MS COCO dataset:
\begin{quotation}
	\noindent
	(a) A young girl \textit{standing on top of a} tennis court. \\
	(b) A giraffe \textit{standing on top of a} green field.
\end{quotation}
The captions describe two very different images. However, comparing these captions using any of the previously mentioned n-gram metrics produces a high similarity score due to the presence of the long 5-gram phrase \textit{`standing on top of a'} in both captions. Now consider the captions (c,d) obtained from the same image:
\begin{quotation}
	\noindent
	(c) A shiny metal pot filled with some diced veggies. \\
	(d) The pan on the stove has chopped vegetables in it.
\end{quotation}
These captions convey almost the same meaning, but exhibit low n-gram similarity as they have no words in common. 

To overcome the limitations of existing n-gram based automatic evaluation metrics, in this work we hypothesize that \emph{semantic propositional content is an important component of human caption evaluation}. That is, given an image with the caption `A young girl standing on top of a tennis court', we expect that a human evaluator might consider the truth value of each of the semantic propositions contained therein---such as (1) there is a girl, (2) girl is young, (3) girl is standing, (4) there is a court, (5) court is tennis, and (6) girl is on top of court. If each of these propositions is clearly and obviously supported by the image, we would expect the caption to be considered acceptable, and scored accordingly.

Taking this main idea as motivation, we estimate caption quality by transforming both candidate and reference captions into a graph-based semantic representation called a \textit{scene graph}. The scene graph explicitly encodes the objects, attributes and relationships found in image captions, abstracting away most of the lexical and syntactic idiosyncrasies of natural language in the process. Recent work has demonstrated scene graphs to be a highly effective representation for performing complex image retrieval queries \cite{Johnson2015,Schuster2015}, and we demonstrate similar advantages when using this representation for caption evaluation.

To parse an image caption into a scene graph, we use a two-stage approach similar to previous works \cite{Schuster2015,Wang2015,Lin2014visual}. In the first stage, syntactic dependencies between words in the caption are established using a dependency parser \cite{Klein2003} pre-trained on a large dataset. An example of the resulting dependency syntax tree, using Universal Dependency relations \cite{de2014universal}, is shown in Figure~\ref{fig:motivation} top. In the second stage, we map from dependency trees to scene graphs using a rule-based system \cite{Schuster2015}. Given candidate and reference scene graphs, our metric computes an F-score defined over the conjunction of logical tuples representing semantic propositions in the scene graph (e.g., Figure~\ref{fig:motivation} right). We dub this approach SPICE for \emph{Semantic Propositional Image Caption Evaluation}.

Using a range of datasets and human evaluations, we show that SPICE outperforms existing n-gram metrics in terms of agreement with human evaluations of model-generated captions, while offering scope for further improvements to the extent that semantic parsing techniques continue to improve. We make code available from the project page\footnote{http://panderson.me/spice}. Our main contributions are: 

\begin{enumerate}
	\item We propose SPICE, a principled metric for automatic image caption evaluation that compares semantic propositional content;
	\item We show that SPICE outperforms metrics Bleu, METEOR, \mbox{ROUGE-L} and CIDEr in terms of agreement with human evaluations; and 
	\item We demonstrate that SPICE performance can be decomposed to answer questions such as `which caption-generator best understands colors?' and `can caption generators count?'

\end{enumerate}

\section{Background and Related Work}

\subsection{Caption Evaluation Metrics}

There is a considerable amount of work dedicated to the development of metrics that can be used for automatic evaluation of image captions. Typically, these metrics are posed as similarity measures that compare a candidate sentence to a set of reference or ground-truth sentences. Most of the metrics in common use for caption evaluation are based on n-gram matching. Bleu \cite{Papineni2002} is a modified precision metric with a sentence-brevity penalty, calculated as a weighted geometric mean over different length n-grams. METEOR \cite{denkowski2014} uses exact, stem, synonym and paraphrase matches between n-grams to align sentences, before computing a weighted F-score with an alignment fragmentation penalty. ROUGE \cite{Lin2004} is a package of a measures for automatic evaluation of text summaries using F-measures. CIDEr \cite{Vedantam2015} applies term frequency-inverse document frequency (tf-idf) weights to n-grams in the candidate and reference sentences, which are then compared by summing their cosine similarity across n-grams. With the exception of CIDEr, these methods were originally developed for the evaluation of text summaries or machine translations (MT), and were subsequently adopted for image caption evaluation.

Several studies have analyzed the performance of n-gram metrics when used for image caption evaluation, by measuring correlation with human judgments of caption quality. On the PASCAL 1K dataset, Bleu-1 was found to exhibit weak or no correlation (Pearson's $r$ of -0.17 and 0.05)~\cite{Kulkarni2013}. Using the Flickr 8K \cite{Hodosh2013} dataset,  METEOR exhibited moderate correlation (Spearman's $\rho$ of 0.524) outperforming ROUGE SU-4 (0.435), Bleu-smoothed (0.429) and Bleu-1 (0.345) \cite{Elliott2014}. Using the PASCAL-50S and ABSTRACT-50S datasets, CIDEr and METEOR were found to have greater agreement with human consensus than Bleu and ROUGE~\cite{Vedantam2015}.

Within the context of automatic MT evaluation, a number of papers have proposed the use of shallow-semantic information such as semantic role labels (SRLs) \cite{Gimenez2007}. In the MEANT metric \cite{Lo2012}, SRLs are used to try to capture the basic event structure of sentences -- `\textit{who} did \textit{what} to \textit{whom}, \textit{when}, \textit{where} and \textit{why}' \cite{Pradhan2004}. Using this approach, sentence similarity is calculated by first matching semantic frames across sentences by starting with the verbs at their head. However, this approach does not easily transfer to image caption evaluation, as verbs are frequently absent from image captions or not meaningful -- e.g. `a very tall building with a train \textit{sitting} next to it' -- and this can de-rail the matching process. Our work differs from these approaches as we represent sentences using scene graphs, which allow for noun / object matching between captions. Conceptually, the closest work to ours is probably the bag of aggregated semantic tuples (BAST) metric \cite{Ellebracht2015} for image captions. However, this work required the collection of a purpose-built dataset in order to learn to identify Semantic Tuples, and the proposed metric was not evaluated against human judgments or existing metrics.

\subsection{Semantic Graphs}

Scene graphs, or similar semantic structures, have been used in a number of recent works within the context of image and video retrieval systems to improve performance on complex queries \cite{Lin2014visual,Johnson2015,Schuster2015}. Several of these papers have demonstrated that semantic graphs can be parsed from natural language descriptions \cite{Lin2014visual,Schuster2015}. The task of transforming a sentence into its meaning representation has also received considerable attention within the computational linguistics community. Recent work has proposed a common framework for semantic graphs called an abstract meaning representation (AMR) \cite{Banarescu2012}, for which a number of parsers \cite{Carbonell2014,Werling2015robust,Wang2015} and the Smatch evaluation metric \cite{Cai2013} have been developed. However, in initial experiments, we found that AMR representations using Smatch similarity performed poorly as image caption representations. Regardless of the representation used, the use of dependency trees as the starting point for parsing semantic graphs appears to be a common theme \cite{Schuster2015,Wang2015,Lin2014visual}.

\section{SPICE Metric}

Given a candidate caption $c$ and a set of reference captions $S = \{s_{1}, \ldots, s_{m}\}$ associated with an image, our goal is to compute a score that captures the similarity between $c$ and $S$. For the purposes of caption evaluation the image is disregarded, posing caption evaluation as a purely linguistic task similar to machine translation (MT) evaluation. However, because we exploit the semantic structure of scene descriptions and give primacy to nouns, our approach is better suited to evaluating computer generated image captions.

First, we transform both candidate caption and reference captions into an intermediate representation that encodes semantic propositional content. While we are aware that there are other components of linguistic meaning---such as figure-ground relationships---that are almost certainly relevant to caption quality, in this work we focus exclusively on \emph{semantic meaning}. Our choice of semantic representation is the \textit{scene graph}, a general structure consistent with several existing vision datasets \cite{Schuster2015,Johnson2015,Plummer2015flickr30k} and the recently released Visual Genome dataset \cite{krishnavisualgenome}. The scene graph of candidate caption $c$ is denoted by $G(c)$, and the scene graph for the reference captions $S$ is denoted by $G(S)$, formed as the union of scene graphs $G(s_i)$ for each $s_i \in S$ and combining synonymous object nodes. Next we present the semantic parsing step to generate scene graphs from captions.

\subsection{Semantic Parsing---Captions to Scene Graphs}

\begin{figure}[t]
	\vskip 0.2in
	\begin{center}
		\centerline{\includegraphics[width=1.0\columnwidth]{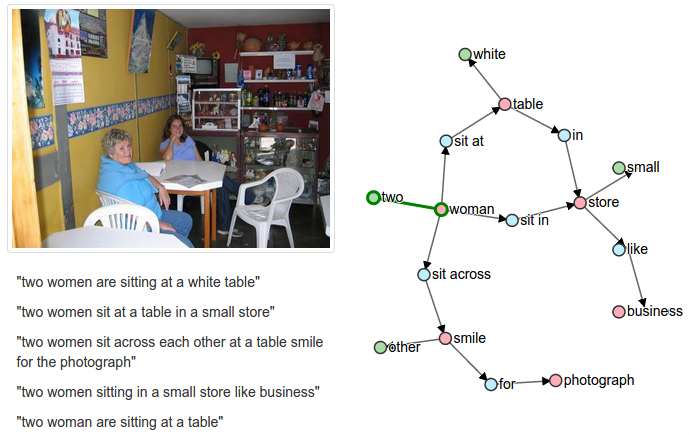}}
		\caption{A typical example of a \textit{scene graph} (right) parsed from a set of reference image captions (left)}
		\label{fig:sg}
	\end{center}
	\vskip -0.2in
\end{figure}

We define the subtask of parsing captions to scene graphs as follows. Given a set of object classes $C$, a set of relation types $R$, a set of attribute types $A$, and a caption $c$, we parse $c$ to a scene graph:
\begin{align}
G(c) &= \langle O(c), E(c), K(c) \rangle
\end{align} 
where $O(c) \subseteq C$ is the set of object mentions in $c$, $E(c) \subseteq O(c) \times R \times O(c)$ is the set of hyper-edges representing relations between objects, and $K(c) \subseteq O(c) \times A$ is the set of attributes associated with objects. Note that in practice, $C$, $R$ and $A$ are \emph{open-world} sets that are expanded as new object, relation and attribute types are identified, placing no restriction on the types of objects, relation and attributes that can be represented, including `stuff' nouns such as grass, sky, etc. An example of a parsed scene graph is illustrated in Figure~\ref{fig:sg}.

Our scene graph implementation departs slightly from previous work in image retrieval \cite{Johnson2015,Schuster2015}, in that we do not represent multiple instances of a single class of object separately in the graph. In previous work, duplication of object instances was necessary to enable scene graphs to be grounded to image regions. In our work, we simply represent object counts as attributes of objects. While this approach does not distinguish collective and distributive readings~\cite{Schuster2015}, it simplifies scene graph alignment and ensures that each incorrect numeric modifier is only counted as a single error.

To complete this subtask, we adopt a variant of the rule-based version of the Stanford Scene Graph Parser~\cite{Schuster2015}. A Probabilistic Context-Free Grammar (PCFG) dependency parser~\cite{Klein2003} is followed by three post-processing steps that simplify quantificational modifiers, resolve pronouns and handle plural nouns. The resulting tree structure is then parsed according to nine simple linguistic rules to extract lemmatized objects, relations and attributes, which together comprise the scene graph. As an example, one of the linguistic rules captures adjectival modifiers, such as the $young \xleftarrow[]{\text{amod}} girl$ example from Figure~\ref{fig:motivation}, which results in the object mention `girl' with attribute `young'. Full details of the pipeline can be found in the original paper.

SPICE slightly modifies the original parser~\cite{Schuster2015} to better evaluate image captions. First, we drop the plural nouns transformation that duplicates individual nodes of the graph according to the value of their numeric modifier. Instead, numeric modifiers are encoded as object attributes. Second, we add an additional linguistic rule that ensures that nouns will always appear as objects in the scene graph---even if no associated relations can identified---as disconnected graph nodes are easily handled by our semantic proposition F-score calculation.

Notwithstanding the use of the Stanford Scene Graph Parser, our proposed SPICE metric is not tied to this particular parsing pipeline. In fact, it is our hope that ongoing advances in syntactic and semantic parsing will allow SPICE to be further improved in future releases. We also note that since SPICE operates on scene graphs, in principle it could be used to evaluate captions on scene graph datasets~\cite{Schuster2015,Johnson2015,Plummer2015flickr30k} that have no reference captions at all. Evaluation of SPICE under these circumstances is left to future work.

\subsection{F-score Calculation}

To evaluate the similarity of candidate and reference scene graphs, we view the semantic relations in the scene graph as a conjunction of logical propositions, or tuples. We define the function $T$ that returns logical tuples from a scene graph as:
\begin{align}
T(G(c)) &\triangleq O(c) \cup E(c) \cup K(c)
\end{align} 
Each tuple contains either one, two or three elements, representing objects, attributes and relations, respectively. For example, the scene graph in Figure~\ref{fig:motivation} would be represented with the following tuples:

\begin{center}
\{ (girl), (court), (girl, young), (girl, standing)\\
(court, tennis), (girl, on-top-of, court) \}
\end{center}

Viewing the semantic propositions in the scene graph as a set of tuples, we define the binary matching operator $\otimes$ as the function that returns matching tuples in two scene graphs. We then define precision $P$, recall $R$, and $SPICE$ as:

\begin{align}
P(c, S) &= \frac{|T(G(c)) \otimes T(G(S))|}{|T(G(c))|}
\\
R(c, S) &= \frac{|T(G(c)) \otimes T(G(S))|}{|T(G(S))|}
\\
SPICE(c, S) &= F_{1}(c, S) = \frac{2 \cdot P(c, S) \cdot R(c, S)}{P(c, S) + R(c,S)}
\end{align}
where for matching tuples, we reuse the wordnet synonym matching approach of METEOR~\cite{denkowski2014}, such that tuples are considered to be matched if their lemmatized word forms are equal---allowing terms with different inflectional forms to match---or if they are found in the same wordnet sysnet. 

Unlike \emph{Smatch}~\cite{Cai2013}, a recently proposed metric for evaluating AMR parsers that considers multiple alignments of AMR graphs, we make no allowance for partial credit when only one element of a tuple is incorrect. In the domain of image captions, many relations (such as \textit{in} and \textit{on}) are so common they arguably deserve no credit when applied to the wrong objects.  

Being an F-score, SPICE is simple to understand, and easily interpretable as it is naturally bounded between 0 and 1. Unlike CIDEr, SPICE does not use cross-dataset statistics---such as corpus word frequencies---and is therefore equally applicable to both small and large datasets.

\subsection{Gameability}

Whenever the focus of research is reduced to a single benchmark number, there are risks of unintended side-effects \cite{Efros2011}. For example, algorithms optimized for performance against a certain metric may produce high scores, while losing sight of the human judgement that the metric was supposed to represent. 

SPICE measures how well caption generators recover objects, attributes and the relations between them. A potential concern then, is that the metric could be `gamed' by generating captions that represent only objects, attributes and relations, while ignoring other important aspects of grammar and syntax. Because SPICE neglects fluency, as with n-gram metrics, it implicitly assuming that captions are well-formed. If this assumption is untrue in a particular application, a fluency metric, such as \textit{surprisal} \cite{hale2001probabilistic,levy2008expectation}, could be included in the evaluation. However, by default we have not included any fluency adjustments as conceptually we favor simpler, more easily interpretable metrics. To model human judgement in a particular task as closely as possible, a carefully tuned ensemble of metrics including SPICE capturing various dimensions of correctness would most likely be the best.

\section{Experiments}

In this section, we compare SPICE to existing caption evaluation metrics. We study both system-level and caption-level correlation with human judgments. Data for the evaluation is drawn from four datasets collected in previous studies, representing a variety of captioning models. Depending on the dataset, human judgments may consist of either pairwise rankings or graded scores, as described further below. 

Our choice of correlation coefficients is consistent with an emerging consensus from the WMT Metrics Shared Task~\cite{wmt15,wmt14} for scoring machine translation metrics. To evaluate system-level correlation, we use the Pearson correlation coefficient. Although Pearson's $\rho$ measures linear association, it is smoother than rank-based correlation coefficients when the number of data points is small and systems have scores that are very close together. For caption-level correlation, we evaluate using Kendall's $\tau$ rank correlation coefficient, which evaluates the similarity of pairwise rankings. Where human judgments consist of graded scores rather than pairwise rankings, we generate pairwise rankings by comparing scores over all pairs in the dataset. In datasets containing multiple independent judgments over the same caption pairs, we also report inter-human correlation. We include further analysis, including additional results, examples and failure cases on our project page\footnote{http://panderson.me/spice}.

\setlength{\tabcolsep}{2.8pt}
\begin{table}[t]
	\centering
	\caption{System-level Pearson's $\rho$ correlation between evaluation metrics and human judgments for the 15 competition entries plus human captions in the 2015 COCO Captioning Challenge~\cite{Chen2015}. SPICE more accurately reflects human judgment overall (M1--M2), and across each dimension of quality (M3--M5, representing correctness, detailedness and saliency) }
	\label{tab:system}
	\begin{tabular}{lcc|cc|cc|cc|cc}
		\hline\noalign{\smallskip}
		& \multicolumn{2}{c}{\textbf{M1}} & \multicolumn{2}{c}{\textbf{M2}} & \multicolumn{2}{c}{\textbf{M3}} & \multicolumn{2}{c}{\textbf{M4}} & \multicolumn{2}{c}{\textbf{M5}} \\
		& $\rho$               & p-value     & $\rho$               & p-value     & $\rho$               & p-value     & $\rho$               & p-value     & $\rho$               & p-value     \\
		\hline\noalign{\smallskip}
		Bleu-1  & 0.24              & (0.369)     & 0.29              & (0.271)     & 0.72              & (0.002)     & -0.54             & (0.030)     & 0.44              & (0.091)     \\
		Bleu-4  & 0.05              & (0.862)     & 0.10              & (0.703)     & 0.58              & (0.018)     & -0.63             & (0.010)     & 0.30              & (0.265)     \\
		ROUGE-L & 0.15              & (0.590)     & 0.20              & (0.469)     & 0.65              & (0.006)     & -0.55             & (0.030)     & 0.38              & (0.142)     \\
		METEOR  & 0.53              & (0.036)     & 0.57              & (0.022)     & 0.86              & (0.000)     & -0.10             & (0.710)     & 0.74              & (0.001)     \\
		CIDEr   & 0.43              & (0.097)     & 0.47              & (0.070)     & 0.81              & (0.000)     & -0.21             & (0.430)     & 0.65              & (0.007)     \\
		SPICE-exact   & 0.84              & (0.000)     & 0.86              & (0.000)     & 0.90              & (0.000)     & 0.39             & (0.000)     & 0.95              & (0.000)     \\
		\textbf{SPICE}   & \textbf{0.88}     & (0.000)     & \textbf{0.89}     & (0.000)     & \textbf{0.89}     & (0.000)     & \textbf{0.46}     & (0.070)     & \textbf{0.97}     & (0.000)    
		\\
		\hline\noalign{\smallskip}  
		M1 & \multicolumn{10}{l}{Percentage of captions evaluated as better or equal to human caption.} \\
		M2 & \multicolumn{10}{l}{Percentage of captions that pass the Turing Test.} \\
		M3 & \multicolumn{10}{l}{Average correctness of the captions on a scale 1--5 (incorrect - correct).} \\
		M4 & \multicolumn{10}{l}{Average detail of the captions from 1--5 (lacking details - very detailed).} \\
		M5 & \multicolumn{10}{l}{Percentage of captions that are similar to human description.}  \\                                                                   
		\hline  
	\end{tabular}
\end{table}
\setlength{\tabcolsep}{1.4pt}

\subsection{Datasets}

\subsubsection{Microsoft COCO 2014.} The COCO dataset \cite{Chen2015} consists of 123,293 images, split into an 82,783 image training set and a 40,504 image validation set. An additional 40,775 images are held out for testing. Images are annotated with five human-generated captions (C5 data), although 5,000 randomly selected test images have 40 captions each (C40 data). 

COCO human judgements were collected using Amazon Mechanical Turk (AMT) for the purpose of evaluating submissions to the 2015 COCO Captioning Challenge~\cite{Chen2015}. A total of 255,000 human judgments were collected, representing three independent answers to five different questions that were posed in relation to the 15 competition entries, plus human and random entries (17 total). The questions capture the dimensions of overall caption quality (M1 - M2), correctness (M3), detailedness (M4), and saliency (M5), as detailed in Table \ref{tab:system}. For pairwise rankings (M1, M2 and M5), each entry was evaluated using the same subset of 1000 images from the C40 test set. All AMT evaluators consisted of US located native speakers, white-listed from previous work. Metric scores for competition entries were obtained from the COCO organizers, using our code to calculate SPICE. The SPICE methodology was fixed before evaluating on COCO. At no stage were we given access to the COCO test captions.  

\subsubsection{Flickr 8K.} The Flickr 8K dataset \cite{Hodosh2013} contains 8,092 images annotated with five human-generated reference captions each. The images were manually selected to focus mainly on people and animals performing actions. The dataset also contains graded human quality scores for 5,822 captions, with scores ranging from 1 (`the selected caption is unrelated to the image') to 4 (`the selected caption describes the image without any errors'). Each caption was scored by three expert human evaluators sourced from a pool of native speakers. All evaluated captions were sourced from the dataset, but association to images was performed using an image retrieval system. In our evaluation we exclude 158 correct image-caption pairs where the candidate caption appears in the reference set. This reduces all correlation scores but does not disproportionately impact any metric.   

\subsubsection{Composite Dataset.} We refer to an additional dataset of 11,985 human judgments over Flickr 8K, Flickr 30K \cite{Young2014} and COCO captions as the composite dataset~\cite{aditya2015images}. In this dataset, captions were scored using AMT on a graded correctness scale from 1 (`The description has no relevance to the image') to 5 (`The description relates perfectly to the image'). Candidate captions were sourced from the human reference captions and two recent captioning models~\cite{Karpathy2015,aditya2015images}.  

\subsubsection{PASCAL-50S} To create the PASCAL-50S dataset \cite{Vedantam2015}, 1,000 images from the UIUC PASCAL Sentence Dataset \cite{Rashtchian2010}---originally containing five captions per image---were annotated with 50 captions each using AMT. The selected images represent 20 classes including people, animals, vehicles and household objects. 

The dataset also includes human judgments over 4,000 candidate sentence pairs. However, unlike in previous studies, AMT workers were not asked to evaluate captions against images. Instead, they were asked to evaluate caption triples by identifying `Which of the sentences, B or C, is more similar to sentence A?', where sentence A is a reference caption, and B and C are candidates. If reference captions vary in quality, this approach may inject more noise into the evaluation process, however the differences between this approach and the previous approaches to human evaluations have not been studied. For each candidate sentence pair (B,C) evaluations were collected against 48 of the 50 possible reference captions. Candidate sentence pairs were generated from both human and model captions, paired in four ways: human-correct (HC), human-incorrect (HI), human-model (HM), and model-model (MM).

\begin{figure*}[t]
	\vskip 0.2in
	\begin{center}
		\includegraphics[width=\linewidth]{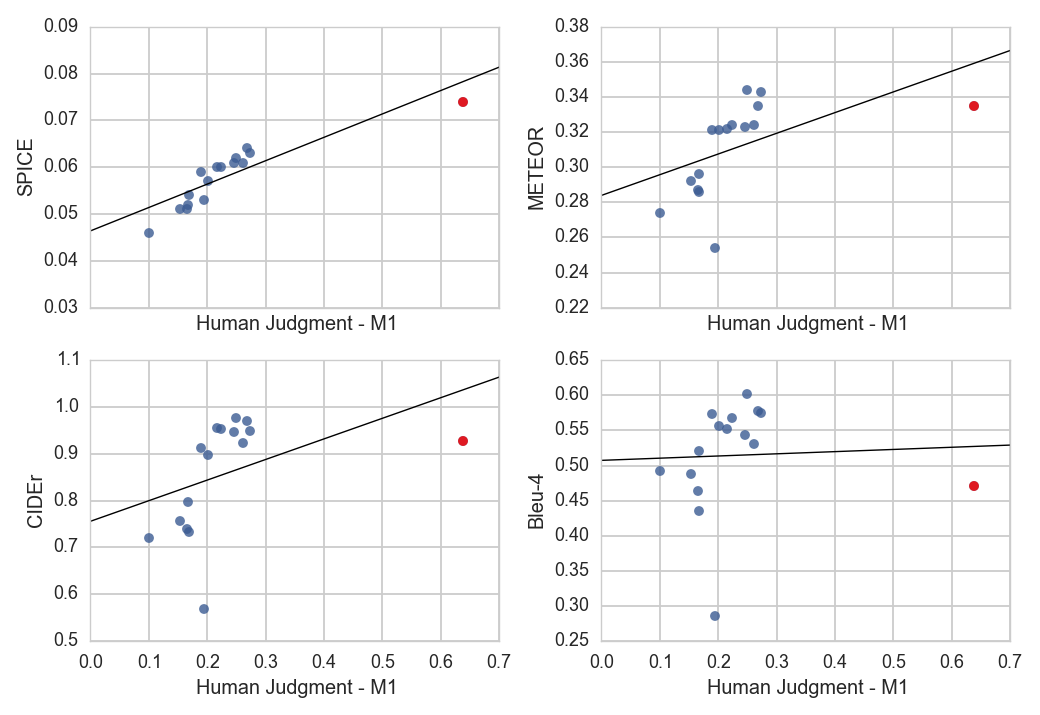}\\
		\caption{Evaluation metrics vs. human judgments for the 15 entries in the 2015 COCO Captioning Challenge. Each data point represents a single model with human-generated captions marked in red. Only SPICE scores human-generated captions significantly higher than challenge entries, which is consistent with human judgment }
		\label{fig:coco_chart}
	\end{center}
	\vskip -0.2in
\end{figure*}

\subsection{System-Level Correlation}
\label{sec:system}

In Table \ref{tab:system} we report system-level correlations between metrics and human judgments over entries in the 2015 COCO Captioning Challenge~\cite{Chen2015}. Each entry is evaluated using the same 1000 image subset of the COCO C40 test set. SPICE significantly outperforms existing metrics, reaching a correlation coefficient of 0.88 with human quality judgments (M1), compared to 0.43 for CIDEr and 0.53 for METEOR. As illustrated in Table \ref{tab:system}, SPICE more accurately reflects human judgment overall (M1 - M2), and across each dimension of quality (M3 - M5, representing correctness, detailedness and saliency). Interestingly, only SPICE rewards caption detail (M4). Bleu and ROUGE-L appear to penalize detailedness, while the results for CIDEr and METEOR are not statistically significant. 

As illustrated in Figure~\ref{fig:coco_chart}, SPICE is the only metric to correctly rank human-generated captions first---CIDEr and METEOR rank human captions 7th and 4th, respectively. SPICE is also the only metric to correctly select the top-5 non-human entries. To help understand the importance of synonym-matching, we also evaluated SPICE using exact-matching only (SPICE-exact in Table \ref{tab:system}). Performance degraded only marginally, although we expect synonym-matching to become more important when fewer reference captions are available.

\subsection{Color Perception, Counting and Other Questions}

Existing n-gram evaluation metrics have little to offer in terms of understanding the relative strengths and weaknesses, or error modes, of various models. However, SPICE has the useful property that it is defined over tuples that are easy to subdivide into meaningful categories. For example, precision, recall and F-scores can be quantified separately for objects, attributes and relations, or analyzed to any arbitrary level of detail by subdividing tuples even further.

To demonstrate this capability, in Table \ref{tab:error} we review the performance of 2015 COCO Captioning Challenge submissions in terms of \textit{color perception}, \textit{counting ability}, and understanding of \textit{size attributes} by using word lists to isolate attribute tuples that contain colors, the numbers from one to ten, and size-related adjectives, respectively. This affords us some insight, for example, into whether caption generators actually understand color, and how good they are at counting. 

As shown in Table \ref{tab:error}, the MSR entry~\cite{Fang2015} ---incorporating specifically trained visual detectors for nouns, verbs and adjectives---exceeds the human F-score baseline for tuples containing color attributes. However, there is less evidence that any of these models have learned to count objects.

\setlength{\tabcolsep}{4pt}
\begin{table*}[t]
	\caption{F-scores by semantic proposition subcategory. SPICE is comprised of object, relation and attribute tuples. Color, count and size are attribute subcategories. Although the best models outperform the human baseline in their use of object color attributes, none of the models exhibits a convincing ability to count
	}
	\label{tab:error}
	\begin{center}
		\begin{tabular}{lccccccc}
			\hline
			& SPICE & Object & Relation & Attribute & Color & Count & Size  \\
			\hline
			Human~\cite{Chen2015}               & \textbf{0.074} & \textbf{0.190}  & \textbf{0.023}    & \textbf{0.054}     & 0.055 & \textbf{0.095}       & \textbf{0.026} \\
			MSR~\cite{Fang2015}                & 0.064 & 0.176  & 0.018    & 0.039     & \textbf{0.063} & 0.033       & 0.019 \\
			Google~\cite{Vinyals2015}              & 0.063 & 0.173  & 0.018    & 0.039     & 0.060 & 0.005       & 0.009 \\
			MSR Captivator~\cite{Devlin2015b}     & 0.062 & 0.174  & 0.019    & 0.032     & 0.054 & 0.008       & 0.009 \\
			Berkeley LRCN~\cite{Donahue2015}       & 0.061 & 0.170  & \textbf{0.023}    & 0.026     & 0.030 & 0.015       & 0.010 \\
			Montreal/Toronto~\cite{Xu2015}    & 0.061 & 0.171  & \textbf{0.023}    & 0.026     & 0.023 & 0.002       & 0.010 \\
			m-RNN~\cite{mao2015learning}               & 0.060 & 0.170  & 0.021    & 0.026     & 0.038 & 0.007       & 0.004 \\
			Nearest Neighbor~\cite{Devlin2015a}    & 0.060 & 0.168  & 0.022    & 0.026     & 0.027 & 0.014       & 0.013 \\
			m-RNN~\cite{mao2014deep} & 0.059 & 0.170  & 0.022    & 0.022     & 0.031 & 0.002       & 0.005 \\
			PicSOM              & 0.057 & 0.162  & 0.018    & 0.027     & 0.025 & 0.000       & 0.012 \\
			MIL                 & 0.054 & 0.157  & 0.017    & 0.023     & 0.036 & 0.007       & 0.009 \\
			Brno University~\cite{KolarHZ15}    & 0.053 & 0.144  & 0.012    & 0.036     & 0.055 & 0.029       & 0.025 \\
			MLBL~\cite{kiros2014multimodal}                & 0.052 & 0.152  & 0.017    & 0.021     & 0.015 & 0.000       & 0.004 \\
			NeuralTalk~\cite{Karpathy2015}         & 0.051 & 0.153  & 0.018    & 0.016     & 0.013 & 0.000       & 0.007 \\
			ACVT                & 0.051 & 0.152  & 0.015    & 0.021     & 0.019 & 0.001       & 0.008 \\
			Tsinghua Bigeye     & 0.046 & 0.138  & 0.013    & 0.017     & 0.017 & 0.000       & 0.009 \\
			Random     & 0.008 & 0.029  & 0.000    & 0.000     & 0.000 & 0.004       & 0.000 \\
			\hline
		\end{tabular}
	\end{center}
\end{table*}
\setlength{\tabcolsep}{1.4pt}

\subsection{Caption-Level Correlation}

In Table \ref{tab:caption-level} we report caption-level correlations between automated metrics and human judgments on Flickr 8K~\cite{Hodosh2013} and the composite dataset~\cite{aditya2015images}. At the caption level, SPICE achieves a rank correlation coefficient of 0.45 with Flickr 8K human scores, compared to 0.44 for CIDEr and 0.42 for METEOR. Relative to the correlation between human scores of 0.73, this represents only a modest improvement over existing metrics. However, as reported in Section \ref{sec:system}, SPICE more closely approximates human judgment when aggregated over more captions. Results are similar on the composite dataset, with SPICE achieving a rank correlation coefficient of 0.39, compared to 0.36 for CIDEr and 0.35 for METEOR. As this dataset only includes one score per image-caption pair, inter-human agreement cannot be established.

\setlength{\tabcolsep}{4pt}
\begin{table}[t]
	\caption{Caption-level Kendall's $\tau$ correlation between evaluation metrics and graded human quality scores. At the caption-level SPICE modestly outperforms existing metrics. All p-values (not shown) are less than 0.001}
	\label{tab:caption-level}
	\begin{center}
		\begin{tabular}{lcccc}
			\hline\noalign{\smallskip}
			& \multicolumn{1}{c}{Flickr 8K \cite{Hodosh2013}} & \multicolumn{1}{c}{Composite \cite{aditya2015images}}  \\
			\hline
			\noalign{\smallskip}
			Bleu-1      & 0.32  & 0.26 \\
			Bleu-4       & 0.14  & 0.18 \\
			ROUGE-L      & 0.32  & 0.28 \\
			METEOR       & 0.42  & 0.35 \\
			CIDEr      & 0.44  & 0.36 \\
			\textbf{SPICE} & \textbf{0.45} & \textbf{0.39} \\
			\hline
			\noalign{\smallskip}
			Inter-human    & 0.73 & - \\
			\hline 
		\end{tabular}
	\end{center}
\end{table}
\setlength{\tabcolsep}{1.4pt}

For consistency with previous evaluations on the PASCAL-50S dataset \cite{Vedantam2015}, instead of reporting rank correlations we evaluate on this dataset using accuracy. A metric is considered accurate if it gives an equal or higher score to the caption in each candidate pair most commonly preferred by human evaluators. To help quantify the impact of reference captions on performance, the number of reference captions available to the metrics is varied from 1 to 48. This approach follows the original work on this dataset \cite{Vedantam2015}, although our results differ slightly which may be due to randomness in the choice of reference caption subsets, or differences in metric implementations (we use the MS COCO evaluation code).

\begin{figure*}[t]
	\vskip 0.2in
	\begin{center}
		\includegraphics[width=\linewidth]{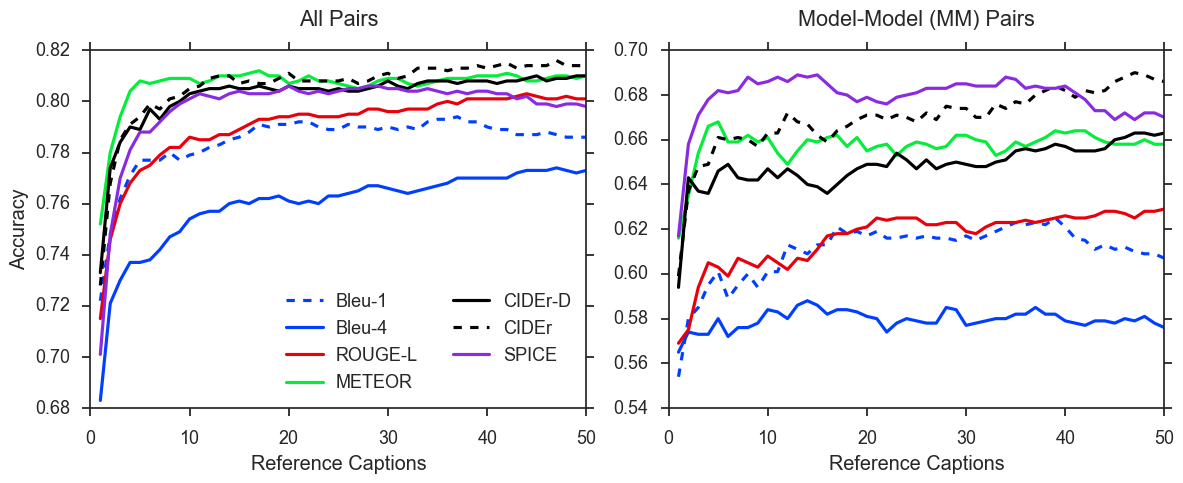}\\
		\caption{Pairwise classification accuracy of automated metrics at matching human judgment with 1-50 reference captions}
		\label{fig:pascal_chart}
	\end{center}
	\vskip -0.2in
\end{figure*}

\setlength{\tabcolsep}{8.0pt}
\begin{table}[t]
	\centering
	\caption{Caption-level classification accuracy of evaluation metrics at matching human judgment on PASCAL-50S with 5 reference captions. SPICE is best at matching human judgments on pairs of model-generated captions (MM). METEOR is best at differentiating human and model captions (HM) and human captions where one is incorrect (HI). Bleu-1 performs best given two correct human captions (HC)}
	\label{tab:pascal_accuracy}
	\begin{tabular}{lccccc}
		\hline\noalign{\smallskip}
		& HC            & HI          & HM           & MM            & All           \\
		\noalign{\smallskip}
		\hline
		\noalign{\smallskip}
		Bleu-1  & \textbf{64.9} & 95.2          & 90.7          & 60.1          & 77.7          \\
		Bleu-2  & 56.6          & 93.0          & 87.2          & 58.0          & 73.7          \\
		ROUGE-L & 61.7          & 95.3          & 91.7          & 60.3          & 77.3          \\
		METEOR  & 64.0          & \textbf{98.1} & \textbf{94.2} & 66.8          & \textbf{80.8} \\
		CIDEr   & 61.9          & 98.0          & 91.0          & 64.6          & 78.9          \\
		SPICE   & 63.3          & 96.3          & 87.5          & \textbf{68.2} & 78.8        \\                                                              
		\hline
	\end{tabular}
\end{table}
\setlength{\tabcolsep}{1.4pt}

On PASCAL-50S, there is little difference in overall performance between SPICE, METEOR and CIDEr, as shown in Figure~\ref{fig:pascal_chart} left. However, of the four kinds of captions pairs, SPICE performs best in terms of distinguishing between two model-generated captions (MM pairs) as illustrated in Table~\ref{tab:pascal_accuracy} and Figure~\ref{fig:pascal_chart} right. This is important as distinguishing better performing algorithms is the primary motivation for this work.

\section{Conclusion and Future Work}

We introduce SPICE, a novel semantic evaluation metric that measures how effectively image captions recover objects, attributes and the relations between them. Our experiments demonstrate that, on natural image captioning datasets, SPICE captures human judgment over model-generated captions better than existing n-gram metrics such as Bleu, METEOR, ROUGE-L and CIDEr. Nevertheless, we are aware that significant challenges still remain in semantic parsing, and hope that the development of more powerful parsers will underpin further improvements to the metric. In future work we hope to use human annotators to establish an upper bound for how closely SPICE approximates human judgments given perfect semantic parsing. We release our code and hope that our work will help in the development of better captioning models.

\small{
	\subsubsection*{Acknowledgements}

	We are grateful to the COCO Consortium (in particular, Matteo R. Ronchi, Tsung-Yi Lin, Yin Cui and Piotr Doll{\'a}r) for agreeing to run our SPICE code against entries in the 2015 COCO Captioning Challenge. We would also like to thank Sebastian Schuster for sharing the Stanford Scene Graph Parser code in advance of public release, Ramakrishna Vedantam and Somak Aditya for sharing their human caption judgments, and Kelvin Xu, Jacob Devlin and Qi Wu for providing model-generated captions for evaluation. This work was funded in part by the Australian Centre for Robotic Vision.
}


\bibliographystyle{splncs}
\bibliography{1022}
\end{document}